\documentclass{article}

\usepackage[final,nonatbib]{nips_2016}

\usepackage[utf8]{inputenc} 
\usepackage[T1]{fontenc}    
\usepackage{url}            
\usepackage{booktabs}       
\usepackage{amsfonts}       
\usepackage{nicefrac}       
\usepackage{microtype}      
\usepackage{graphicx}
\PassOptionsToPackage{square,numbers}{natbib}

\newcommand{\yy}[1]{}

\newcommand{\cd}[1]{{\small \texttt{{#1}}}}

\newcommand{\trm}[1]{\textit{#1}}
\newcommand{\vek}[1]{\textbf{#1}}

\newcommand{\M}{\textbf{M}}

\newcommand{\onenorm}[1]{{|\!|{#1}|\!|_1}}

\title{Neural Query Language:\\A Knowledge Base Query Language for Tensorflow}

\author{
  William W. Cohen\\
    Google Research\\
  \texttt{wcohen@google.com} \\
    \And
  Matthew Siegler \\
      Google Research\\
  \texttt{msiegler@google.com} \\
  \And Alex Hofer\\
      Google\\
  \texttt{rofer@google.com} \\
}

\begin{document}

\maketitle

\begin{abstract}
Large knowledge bases (KBs) are useful for many AI tasks, but are
difficult to integrate into modern gradient-based learning systems.
Here we describe a framework for accessing soft symbolic database
using only differentiable operators. For example, this
framework makes it easy to conveniently write neural models that
adjust confidences associated with facts in a soft KB;
incorporate prior knowledge in the form of hand-coded KB access rules; or learn to instantiate query templates using information extracted from text.
NQL can work well with KBs with millions of tuples and hundreds of thousands of entities on a single GPU.
\end{abstract}

\section{Introduction}

Large knowledge bases (KBs) are useful for many AI tasks, but are
difficult to integrate into modern gradient-based learning systems.
Here we describe the Neural Query Language (NQL),
a framework for accessing soft symbolic databases
using only differentiable operators from Tensorflow \cite{abadi2016tensorflow}. NQL is a dataflow language, implemented in Python and Tensorflow, that provides differentiable operations over (sets of) entities and relations in a KB.
NQL makes it easy to conveniently write neural models that
perform actions that are otherwise difficult.  For instance, a model
can adjust confidences associated with facts in a symbolic KB;
incorporate prior knowledge in the form of hand-coded KB access rules;
learn new KB access rules, thus implementing a variant of
\trm{inductive logic programming}. 

NQL can also be used in a system that will learn to answer natural-language queries against a
KB in a fully end-to-end manner, trained using examples consisting of
a natural-language query input and a entity-set output.  For example, a question like ``who was the father of Queen Victoria's husband?'' might require the following steps to answer: 
\begin{enumerate}
    \item Find the KB entity $e_1$ corresponding to ``Queen Victoria'' in a KB, and find the KB relations $r_1$ and $r_2$ that correspond the ``husband'' and ``father of''.
    \item Use the KB to find the entity or entities $e_2$ that are related to $e_1$ via the relationship $r_1$, and then find the entity or entities $e_3$ that are related to $e_1$ via the relationship $r_3$.
\end{enumerate}
Neural networks can be trained to perform the first step above: finding $e_1$ is an entity-linking task and finding $r_1$ and $r_2$ is a relation extraction task.  Using NQL, the second step can also be performed with differentiable operators.  This means that the loss between the predicted answer (in this case $e_3$) and the desired answer can be backpropagated all the way to the entity-linking and relation extraction networks.

\section{Related Work}

NQL is closely related to TensorLog \cite{cohentensorlog}, a deductive database formalism which also can be compiled to Tensorflow.  In fact, NQL was designed so that every expression in the target sublanguage used by TensorLog can be concisely and
readably written in NQL.  TensorLog, in turn, has semantics derived from other ``proof-counting'' logics such as stochastic logic programs (SLP) \cite{DBLP:journals/ml/Cussens01}. TensorLog is also closely related to other differentiable first-order logics  such as the differentiable theorem prover (DTP) \cite{rocktaschel2016learning}, in which a proof for an example is unrolled into a
network.  DPT includes representation-learning as a
component, as well as a template-instantiation approach similar to the one used in NQL.
TensorLog and NQL are more restricted than DPT but also more scaleable: the current version of NQL can work well with KBs with millions of tuples and hundreds of thousands of entities, even on a single GPU.  

NQL however is not a logic, like TensorLog, but a dataflow language, similar in spirit to Pig \cite{gates2009building} or Spark \cite{zaharia2010spark}.  NQL also includes a number of features not found in TensorLog, notably the ability to have variables that refer to relations.  NQL also makes it much easier for Tensorflow models to include pieces of NQL, or for NQL queries to call out to Tensorflow models.

NQL is one of many systems that have been built on top of Tensorlog or some other deep-learning platform.  Perhaps the most similar of these in spirit is Edward \cite{tran2016edward}, which like NQL, attempts to add a higher-level modeling language based on a rather different programming paradigm: most other packages are aimed at providing additional support for training, or combining existing Tensorflow operators into reusable fragments.  In the case of Edward, the alternative paradigm being supported is probabilistic programming (e.g., variational autoencoder modes), while in Tensorlog, the alternative paradigm supported is dataflow operations on KGs.

\section{NQL: A Neural Query Language}

\subsection{Preliminaries}

\newcommand{\ty}{\textit{type}}
\newcommand{\nam}{\textit{name}}
\newcommand{\aty}{\tau}
\newcommand{\myR}{\mathbb{R}}

NQL allows one to query a KB of \trm{entities} and
\trm{relations}.  An \trm{(typed) entity} $e$ has a type $\ty(e)$, and an
index $i(e)$, which is an integer between $1$ and $N_{\ty(e)}$, where
$N_\aty$ is the number of entities of type $\aty$.
Types and entities both have \trm{names}, which are
readable strings describing them: the name of an entity $e$, for
instance, will be written $\nam(e)$ below.
We assume that
names and indices for entities are unique within a type, so if
$\ty(e)=\ty(e')$ and either $i(e)=i(e')$ or $\nam(e)=\nam(e')$, then it must
be that $e=e'$.  

A \trm{weighted relation} $\pi$ with \trm{domain type} $\aty_1$ and
\trm{range type} $\aty_2$ is a weighted multiset of pairs of entities
$(e_1,e_2)$ such that $\ty(e_1)=\aty_1$ and $\ty(e_2)=\aty_2$.
NQL currently supports only binary relations.
Relations can be thought of as weighted edges from nodes of type
$\aty_1$ to nodes of type $\aty_2$.  A weighted relation $\pi$ can be
encoded as a (possibly sparse) matrix $\vek{M}_\pi \in \myR^{N_{\aty_1}
  \times N_{\aty_2}}$.  Relations also have string names.

A KB is a pair $(\Pi,E)$ where $\Pi=\pi_1,\ldots,\pi_{N_\Pi}$ is a set
of relations, and $E$ is a set of typed entities. 

NQL also makes use of weighted multisets of typed entities.  A
\trm{weighted multiset} $\sigma$ of type $\aty$ is a mapping from
entities of type $\aty$ to non-negative real numbers, which we will
write in a Python-like notation, e.g. \cd{\{blue:0.9, red:1.0\}}.
Entities of type $\aty$ not explicitly listed in this notation are
assumed to map to zero.  A weighted multiset $\sigma$ of type $\aty$ can be
encoded as a (possibly sparse) vector $\vek{v}_\sigma \in
\myR^{N_\aty}$, where $\vek{v}_\sigma[i(e)]=\sigma(e)$.

\subsection{Simple NQL expressions}

\begin{table}[t]
\begin{center}
\begin{tabular}{rcll}
NQL expression & & Vector-matrix specification & Comments\\ 
\hline
\cd{s.rel()}  & $\equiv$ & $\vek{s} \M_\pi$ \\
\cd{s.rel(-1)}  & $\equiv$ & $\vek{s} \M_\pi^T$ \\
\cd{s | t}  & $\equiv$ & $\vek{s} + \vek{t}$  \\
\cd{s \& t}  & $\equiv$ & $\vek{s} \odot \vek{t}$  & $\odot$ \textit{is Hadamard product}\\
\cd{s.follow(r)}  & $\equiv$ & $\vek{s} \left( \sum_{i=1}^k \vek{r}[i] M_{\pi_i} \right)$ \\
\cd{s.if\_any(t)}  & $\equiv$ & $\vek{s} \onenorm{\vek{t}}$ \\
\cd{s * a}  & $\equiv$ & $\vek{s} a$ & $a$ is a Tensorflow scalar\\
\hline
\end{tabular}
\end{center}

\caption{Matrix-vector implementation for NQL operators.
Vectors $\vek{s}, \vek{t}, \vek{r}$ correspond to \cd{s, t, r} respectively,
$\vek{r}$ is over the relation group $\pi_1,\ldots,\pi_k$,
  and $\M_\pi$ corresponds to the relation \cd{rel}.}
\label{tab:nql}
\end{table}

NQL is a simple KB query language embedded in Python.  Some NQL
expressions are produced using an \trm{NQL context object}, which
contains pointers to a KB.  Below I will use the variable \cd{c} for
an NQL context object, and assume it has been initialized with a
database derived from a widely-used example database of geneology
information about European royal families\footnote{The dataset is
  widely distributed as an example of the GED format, for example
  under https://github.com/jdfekete/geneaquilt} from which we have
derived 12 familial relations named \cd{aunt}, \cd{brother},
\cd{daughter}, \cd{father}, \cd{husband}, \cd{mother}, \cd{nephew},
\cd{niece}, \cd{sister}, \cd{son}, \cd{uncle}, and \cd{wife}.  This KB
has only one type, \cd{person\_t}, and all the relations/edges have
unit weight.

\begin{figure}[b]
\centerline{\includegraphics[width=\linewidth]{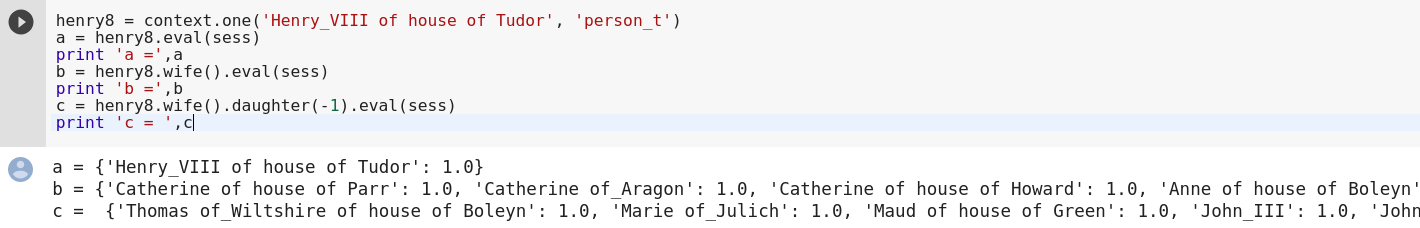}}
\caption{Example: Using NQL to compute the six wives (b) and twelve
  in-laws (c) of King Henry VIII (a).  The variable \cd{sess} is bound
  to a Tensorflow \cd{Session} object.} \label{fig:nql}
\end{figure}

One can create singleton, unit-weighted sets using a call to
\cd{c.one}, for example:

\cd{~~henry8 = c.one(`Henry\_VIII of house of Tudor', 'person\_t')}

Evaluating and printing the NQL expression \cd{henry8} would yield the
multiset \cd{\{'Henry\_VIII of house of Tudor': 1.0\}}.  There are two
other primitive set-constructions methods for contexts, \cd{c.none},
which creates an empty set of a given type, and \cd{c.all}, which
creates a universal unit-weighted set of a given type.

Every relation $\pi$ can be accessed by simply using the name of that
relation as a method of any multiset-valued expression.  For instance
\cd{henry8.wife()} would evaluate to a set of six people, \cd{\{`Anne
  of house of Boleyn':1.0, `Anne of\_Cleves':1.0,}, \ldots, \cd{`Jane
  of house of Seymour':1.0\}}.  Relations can also be chained: for
example, the set of sons of Henry VIII's daughters could be
written as \cd{henry8.daughter().son()}.  

One can also reference a relation by a string ``\cd{r}'' that names it
with the syntax \cd{s.follow(`r')}: for instance
\cd{henry8.follow(`wife')} gives the same set as above.  The inverse
of a relation can by accessed by adding an argument \cd{-1} to a
relation-name method: for example, Henry VIII's parents could be
found with the expression \cd{henry8.son(-1)}.

Unions and intersections of multisets of the same type can be computed
using the operators ``\cd{|}'' for union and ``\cd{\&}'' for
intersection.  For example, the set of henry VIII's grandsons
could be written as

\cd{~~(henry8.son() | henry8().daughter()).son()}

Since the language is embedded in Python, one can use Python's
function definition construct to define NQL functions.  As an example,
this definition

\cd{~~def child(x): return son(x) | daughter(x)}

would allow one to re-express the set defined above as
\cd{child(henry8).son()}.  (One can also define new multiset
methods, so that the notation \cd{henry8()} could be used,
by subclassing the NQL context class.)

\subsection{Conditionals, Predicate Variables, and Rule Templates}

NQL also has a conditional construct.  If \cd{s} and \cd{t} are
multisets then \cd{s.if\_any(t)} returns exactly the set \cd{s} if
\cd{t} is a singleton multiset with weight one on its only element,
and returns the empty set of \cd{t} is an empty set.  More generally,
\cd{s.if\_any(t)} will return a copy of \cd{s} in which every element
has been multiplied by a factor $f$, where $f$ is the sum of all the
weights of the members of \cd{t}.  This operation is best described
using the vectors $\vek{s}$ and $\vek{t}$ corresponding to \cd{s} and
\cd{t} respectively: the vector $\vek{s}'$ corresponding to
\cd{s.if\_any(t)} is simply \( \vek{s}' \equiv \vek{s}
\onenorm{\vek{t}} \).  This definition, along with the definitions of
the other NQL operators, is shown in Table~\ref{tab:nql}.  It is also
possible to obtain a similar conditional effect with the notation
\cd{s * a} where \cd{a} is a Tensorflow scalar.

NQL also includes a construct which allows one to construct variables
which range over relations.  Any set of relations $\pi_1,\ldots,\pi_k$
with the same domain and range types can be gathered together into
\trm{relation group} $g$.  This creates a new type $\tau_g$ with
$N_{\tau_g}=k$ elements, whose entity members have the same names as
the relations $\pi_1,\ldots,\pi_k$.  One can then use the same NQL
constructs to create weighted multisets of relations.

For example, the \cd{rel\_t} is a type for all of the relations in
this KB, one could create the multiset 

\cd{~~child = c.one('daughter', 'rel\_t') | c.one('son', 'rel\_t')}

If \cd{r} is a multiset of relation-naming entities, then the syntax
\cd{s.follow(r)} also lets you ``follow'' a group of relations.  So in
this example, \cd{henry8.follow(child)} would evaluate to the set
of all daughters and sons of Henry VIII.  More generally, the
weights associated with each relation in \cd{r} are combined
multiplicatively with any weights associated with the edges in the KB
itself: a definition for this operator is also shown in
Table~\ref{tab:nql}.

\section{Learning and Rule Templates}

\subsection{NQL and Tensorflow}

NQL is tightly integrated with Tensorflow.
Every NQL expression is attached to a context object \cd{c}.  The context object has sufficient information to produce an appropriate Tensorflow compute-graph node which is an implementation of the NQL expression.  These Tensorflow expressions are computed bottom-up.  

If \cd{x} is an NQL expression, one can access the underlying Tensorflow implementation using the syntax \cd{x.tf}.  If $\tau$ names an NQL type and \cd{c} is an NQL context, and if \cd{w} is a compatible Tensorflow \cd{Tensor} or \cd{Variable} object, then \cd{c.as\_nql(w, $\tau$)} converts \cd{w} to an NQL expression.  (By ``compatible'' here we mean that \cd{w} contains a tensor of the right shape, i.e., it contains a minibatch of vectors in $\myR^{N_\tau}$.) This makes it relatively simple to convert back and forth between NQL and Tensorflow, so models can easily include Tensorflow submodels (e.g., an LSTM to encode represent text) as well as NQL templates.

The current implementation of NQL can handle KBs with a few million tuples and types with a few hundred thousand entities on a single commodity GPU.

\subsection{Learning with NQL} \label{learning-with-nql}

Having variables that can be bound to multisets of relations makes it
possible to write relatively generic ``template'' queries.  For
instance, in the family-relations domain, many of the relations can be approximated
with the union of a small number of queries that each chain together two
other relations: e.g., \cd{x.father()} is approximately the same as
\cd{x.mother().husband()}, and \cd{x.daughter()} is approximately the
same as \cd{x.daughter().sister() | x.son().sister()}.  

If one wanted to learn to approximate a new familial relation with the
ones in this dataset, one might pose the following learning problem:
learn values for the multiset relation variables \cd{r1}, \cd{r2},
\cd{r3} and \cd{r4} for the query:

\cd{~~x.follow(r1).follow(r2) | x.follow(r3).follow(r4)}

This ``template'' could be turned into an approximation of \cd{father}
by setting \cd{r1=r3=\{mother:1.0\}}, \cd{r2=r4=\{husband:1.0\}}, or
into an approximation of \cd{daughter} by setting
\cd{r1=\{daughter:1.0\},r3=\{son:1.0\}}, \cd{r2=r4=\{sister:1.0\}}.
Templates reduce the difficult problem of searching a discrete space
of possible queries to the more tractable problem of searching a
continuous space of weights inside a multiset.  

\begin{table}[t]
\begin{tabular}{l}
\hline
\textit{Input: an entity $x$; Output: entity-set $y$ so that $y=\{ y':\pi(x,y) \}$}\\
\cd{~~def trainable\_rel\_var():}\\
\cd{~~~~return c.as\_nql(tf.Variable(tf.ones\_initializer()[k]))} \\
\cd{~~r1 = trainable\_rel\_var()} \\
\cd{~~r2 = trainable\_rel\_var()} \\
\cd{~~r3 = trainable\_rel\_var()} \\
\cd{~~r4 = trainable\_rel\_var()} \\
\cd{~~y = x.follow(r1).follow(r2) | x.follow(r3).follow(r4)}\\
\cd{~~loss = $\ell(\cd{y.tf},\cd{target\_labels})$}\\
\hline
\textit{Inputs: a question $q$ containing an entity \cd{e}; Output: entity-set $y$ answering the question $q$.}\\  
\cd{~~rel = c.as\_nql($f(\cd{q})$)}\\
\cd{~~y = e.follow(rel)} \\
\cd{~~loss = $\ell(\cd{y.tf},\cd{target\_labels})$}\\
\hline
\textit{Inputs: a question $q$ containing an entity \cd{e}; Output: entity-set $y$ answering the question $q$.}\\  
\cd{~~r1 = c.as\_nql($f_1(\cd{q})$)}\\
\cd{~~r2 = c.as\_nql($f_2(\cd{q})$)}\\
\cd{~~switch1 = $f_3(\cd{q})$}\\
\cd{~~switch2 = $f_4(\cd{q})$}\\
\cd{~~y = e.follow(r1) * switch1 | e.follow(r1).follow(r2) * switch2} \\
\cd{~~loss = $\ell(\cd{y.tf},\cd{target\_labels})$}\\
\hline
\end{tabular}
\caption{Some example learning based on NQL templates.  In each of these the $f$'s are differentiable functions of a textual query $q$, e.g., based on an encoder-decoder approach.  }  \label{tab:models}
\end{table}

Using this template, a very minimal approach to learning rules for an
unknown predicate $\pi_*$ would be the following.  (We assume the
rules will take as input a domain entity \cd{x} and output a set of
things related to \cd{x} via $\pi_*$.)  First, let \cd{r1}, \ldots,
\cd{r4} be NQL relation variables derived from Tensorflow \cd{Variable}s
with shape $k$, where $k$ is the number of relations.  Second, let the models
prediction \cd{y} be defined using the template above, i.e., let

\cd{~~y = x.follow(r1).follow(r2) | x.follow(r3).follow(r4)}

Third, define an appropriate loss function on \cd{y} and train.  This
learns a definition of the predicate in terms of the values of
\cd{r1}, \ldots, \cd{r4}.  This approach is summarized on the top of Table~\ref{tab:models}.

Another example use of templates is for question-answering against a
KB.  For example, consider simple questions of the form ``Who was the
father of Queen Victoria?'' which ask for entities in some particular
relation (e.g., \cd{father}) to a specific ``seed entity'' \cd{e}
appearing in the question (e.g., \cd{'Victoria of house of Hanover'}).
If there is an entity-linking system that can extract the appropriate
entity \cd{e} from a question $q$, then a simple question-answering
system can be defined using the model in the second panel of
Table~\ref{tab:models}.  Here $f$ would be an arbitrary
differentiable function of $q$, e.g., based on decoding an LSTM to the
relation variable.

Clearly, these approaches could be combined to consruct models for
more complex, multi-hop, compositional queries, like ``who was the
father of Queen Victoria's husband?'', as shown in the bottom panel of
Table~\ref{tab:models}.  Here $f_1\ldots f_4$ are based on decoding
$q$, and \cd{switch1} and \cd{switch2} are ``switches'' which select
whether a one-hop or two-hop query was selected.

\subsection{Learning without Templates}

\yy{todo: Alex - maybe 1 pg on how to work with more flexibly-defined space of rules, eg consider a classifier for starting point and variable length seq of follow operations.
could encode sentence, produce starting point, then decode sentence for k steps and ...}

\begin{table}[t]
\begin{tabular}{l}
\hline
\textit{Input: an initial entity $e$ and encoded state $s$; Output: entity-set $y$ }\\
\cd{~~p = 1; y = tf.zeros(k)}\\
\cd{~~for i in range(MAX\_HOPS):}\\
\cd{~~~~s, r, p\_stop = $f$(s)}\\
\cd{~~~~e = e.follow(r)}\\
\cd{~~~~y += p * p\_stop * e.as\_tf()}\\
\cd{~~~~p = p * (1 - p\_stop)}\\
\hline
\end{tabular}
\caption{An example learning using NQL without templates.  Here $f$ is a recurrent, differentiable function of a state vector $s$ which
returns a new state vector $s$, a distribution over relations $r$, and a probability of stopping $p$. }  \label{tab:templatefree}
\end{table}

Consider the question answering task described at the end of Section
\ref{learning-with-nql} with a given seed entity $e$ and potentially
multiple relations necessary to reach the target. As shown above
this task can be learned using templates. However, this can 
grow unwieldy as the depth of reasoning increases.

Instead, one could take advantage of the structure inherent in these
templates and build a model which learns over an entire family of templates.
Each of the templates mentioned in the question answering task represents a
series of relations followed sequentially where the relations to follow and
the order to follow them in are what the model is learning.

We can use a recurrent model to predict any chain of relations up to an
arbitrary length. Consider a recurrent model
$s_i, r_i, p_i = f(s_{i-1})$ where $s_i$ is the state at step $i$,
$r_i$ is a predicted relation for step $i$, $p_i$ is the probability
of stopping at step $i$, and $s_0$ is an encoding of the query.

Using this model a final prediction can be calculated by weighting predictions from all number of steps up
to some maximum as shown in Table~\ref{tab:templatefree}.

Representing the problem in this way has the advantage that it allows
for generalization to questions which require deeper reasoning than
seen at training time. For example, a model trained on data which requires
following up to $5$ relations may successfully return answers requiring 
$6$ or more relations to be followed.

This approach may be extended further to cover other more complex families
of templates.

\yy{summarized in table X.}

\section{Advantages and Disadvantages of NQL}

\subsection{NQL \textit{vs.}~TensorLog}

NQL's implementation is fairly thin: it is implemented by having NQL
expressions converted directly to Tensorflow computation graphs.
During this conversion process NQL also enforces type checking (e.g.,
to ensure that the $\aty$'s for the domain and range of each relation
is consistent). The \cd{eval} method for NQL expressions makes use of
backpointers to the context to allow conversion to the symbolic names
of entities.

As Table~\ref{tab:nql} shows, NQL's operations can be concisely
specified with matrix-vector computations: it might be asked how much
additional value the NQL abstraction provides over Tensorflow.  We
should note that the actual implementation of NQL's operators are less
concise than the specification, for a number of reasons.\footnote{For
  instance, for even a small KB, it is essential that the $\M_\pi$
  matrices are stored as \textit{sparse} matrices, but currently
  Tensorflow does not support sparse-sparse matrix multiplication, and
  multiplication of sparse matrix to a dense matrix is only supported
  in one order.}  There are also a number of plausible ways to
implement the \cd{s.follow(r)} operation, and using this higher-level
notation allows one to choose between them easily as a configuration
option.

\subsection{NQL \textit{vs.}~SQL (or SPARQL or OWL or ....)}

Compared to more traditional KB query languages, NQL has a major
limitation, in that it cannot construct and return tuples, only
weighted sets of entities.  This is a direct consequence of the
decision to base NQL on differentiable vector-matrix operations, which
do not support creating new objects (such as tuples).

This limitation seems to make it impossible for NQL to perform a
number of familiar DB operations, such as joins.  Consider two tables
\cd{student} and \cd{grade}, \cd{student} having fields \cd{id},
\cd{program}, and \cd{expected\_degree} and \cd{grade} having fields
\cd{student\_id}, \cd{course\_id}, \cd{letter\_grade}.  Consider an
SQL join query like

\cd{SELECT grade.id FROM student, grade}\\
\cd{WHERE student.id = grade.student\_id}\\
\cd{AND student.expected\_degree = 'PhD' AND grade.letter\_grade = 'C'}

which asks for PhD students that have gotten a C in some course.
Although this seems to need more power than NQL has, it can be
emulated by incorporating into the KB new structures which act as
indexes for the relations.  In this case, one could construct a
\cd{student\_record} type and a \cd{grade\_record} type, and define
relations such as \cd{student\_record\_id}, mapping
\cd{student\_record} entities to the appropriate \cd{id} value, and so
on.  The SQL query above could be emulated with the NQL code

\cd{c\_records = c.one('C', 'letter\_grade\_t').grade\_record\_letter\_grade(-1)}\\
\cd{records\_of\_students\_with\_Cs = c\_records.grade\_record\_student\_id().student\_record\_id(-1)}\\
\cd{records\_of\_phds = c.one('PhD', 'degree\_t').student\_record\_expected\_degree(-1)}\\
\cd{result = (records\_of\_students\_with\_Cs \& records\_of\_phds).student\_record\_id}

Although many join-like queries can be treated this way, SQL queries
that return novel tuples clearly cannot be performed in NQL (e.g., if
we modified the query above to \cd{SELECT} both a course id and a
student id.)  However, NQL has an advantage over more expressive query
languages in that it is differentiable, so it is possible to base a
differentiable loss function on the result of executing an NQL query.

\section{Conclusion}

We have described NQL, a query language which makes it convenient to integrate queries on a KB into a neural model implemented in Tensorflow.  NQL accesses data using only differentiable operators from Tensorflow, which allows a tight integration with gradient-based learning methods.  NQL is available as open source: because of its many applications in NLP, code for NQL is available at \cd{https://github.com/google-research/language}, the Google Research repository for NLP components.


\bibliographystyle{plain}
\bibliography{./all}  

\end{document}